\def\year{2021}\relax
\documentclass[letterpaper]{article} 
\usepackage{aaai21}  
\usepackage{times}  
\usepackage{helvet} 
\usepackage{courier}  
\usepackage[hyphens]{url}  
\usepackage{graphicx} 
\usepackage{natbib}  
\usepackage{caption} 
\usepackage{amsmath,amssymb,amsfonts}
\usepackage{multirow}
\usepackage{makecell}
\usepackage{hhline}

\usepackage[utf8]{inputenc}
\usepackage[dvipsnames]{xcolor}

\title{Bidirectional RNN-based Few Shot Learning for 3D Medical Image Segmentation}
\author{
    Soopil Kim \textsuperscript{\rm 1}, Sion An\textsuperscript{\rm 1},  Philip Chikontwe\textsuperscript{\rm 1}, Sang Hyun Park\textsuperscript{\rm 1}
    \\
}
\affiliations{
    \textsuperscript{\rm 1}Department of Robotics Engineering, DGIST\\
soopilkim, sion\_an, philipchicco, shpark13135@dgist.ac.kr
}

\begin{document}

\maketitle

\begin{abstract}
Segmentation of organs of interest in 3D medical images is necessary for accurate diagnosis and longitudinal studies. Though recent advances using deep learning have shown success for many segmentation tasks, large datasets are required for high performance and the annotation process is both time consuming and labor intensive. In this paper, we propose a 3D few shot segmentation framework for accurate organ segmentation using limited training samples of the target organ annotation. To achieve this, a U-Net like network is designed to predict segmentation by learning the relationship between 2D slices of support data and a query image, including a bidirectional gated recurrent unit (GRU) that learns consistency of encoded features between adjacent slices. Also, we introduce a transfer learning method to adapt the characteristics of the target image and organ by updating the model before testing with arbitrary support and query data sampled from the support data. We evaluate our proposed model using three 3D CT datasets with annotations of different organs. Our model yielded significantly improved performance over state-of-the-art few shot segmentation models and was comparable to a fully supervised model trained with more target training data.
\end{abstract}

\section{Introduction}
\noindent Deep learning based segmentation models have achieved success in various applications \cite{gu2019net, zhou2018unet, alom2019recurrent, nie2019difficulty} for both natural and medical images. Despite their success, it is often difficult to make a robust model for segmentation especially for medical images since constructing a large-scale dataset incurs a high cost in scanning, and manually creating annotations for volumetric images is laborious and time-consuming. Above all, each hospital acquires images with different resolutions and modalities given that medical experts are interested in different tasks. Consequently, it would constitute designing a separate model for each task, which is not practical. Moreover, each task may have a low data regime with limited annotated samples, and training based on fine-tuning or transfer learning may fail and lead to overfitting.

Recently, few shot learning methods \cite{snell2017prototypical,sung2018learning,garcia2017few,liu2018learning} have been proposed in the field of machine learning to efficiently address these challenges. The key idea of few shot learning is to learn generalizable knowledge across several related tasks that can be used to predict the label of a query sample with support data and labels. Therefore, in the ideal setting, when this idea is applied to medical datasets, a model trained with several existing organ annotations should be able to accurately segment unseen target organs with only a few samples. However, most few shot methods that focus on natural 2D images are not directly applicable for 3D image based analysis since such models are prone to overfitting when trained with few observations. 

In general, 3D operations  for dense pixel-level high dimensional predictions  incur increased high memory usage and often lead to constraining the batch size to be small. Recently, \cite{roy2020squeeze} proposed a framework for organ segmentation in 3D CT scans using few shot learning. Volumetric segmentation is performed in 2D slice by slice in the axial view with careful selection of potential query and support slices. Though successful, their approach does not consider the relation between adjacent slices and 3D structural information, therefore the segmentation result may often be inaccurate and not smooth.

In this paper, we propose a novel few shot segmentation framework that models the relation between support and query data from other few shot tasks alongside 3D structural information between adjacent slices. We integrate a bidirectional gated recurrent unit (GRU) between the encoder and decoder of a 2D few shot segmentation model for efficient representation learning. In this way, encoded features of both the support set and adjacent slices capture key characteristics to predict the segmentation of a query image in the decoding layers. Furthermore, we propose a transfer learning strategy to adapt the characteristics of a target domain in multi-shot segmentation setting. For a given task, we re-train the model parameters using the given support data by arbitrarily dividing them into support and query data using data augmentation. We evaluated our method using 3 datasets (1 for internal test and the rest for external validation) to verify the generalization ability of the model. 

In our experiments, the proposed model using less than 5 support samples could achieve performance comparable to a supervised learning method used a large number of data samples. The contribution points of this paper are summarized as follows:
\begin{itemize}
\item We propose a novel 3D few shot segmentation model able to capture key relationships between adjacent slices in volume via bidirectional GRU modules.
\item We propose a transfer learning strategy to improve performance in multi-shot segmentation models. 
\item We empirically demonstrate the generalization ability of our method through several experiments on both internal and external datasets with various organs and different resolutions.  
\end{itemize}

\section{Related Works}
\subsection{Segmentation using Few Shot Learning}
Few shot learning aims to learn transferable knowledge that can be used to generalize to unseen tasks with scarce labeled training data. It has been widely used for classification \cite{snell2017prototypical,sung2018learning,garcia2017few,liu2018learning}, segmentation \cite{li2020fss,wang2019panet,zhang2019canet,shaban2017one,rakelly2018conditional}, and regression tasks \cite{finn2017model,wang2017learning,zhou2019efficient}. For the segmentation task, \cite{shaban2017one} first proposed a method with a fully convolutional network (FCN) that learns optimal parameters from a support image and its label to segment the relevant object from in a query image. \cite{dong2018few} defined a prototype, i.e., a feature vector with high-level discriminative information, through encoder and global average pooling from support data, then predicted segmentation based on similarity with the prototype of the query image. \cite{wang2019panet} additionally proposed an alignment loss that inversely guesses the support label by using the query image and the predicted query label as support data in order to select a more accurate prototype. \cite{zhang2019canet} proposed a model that refines the segmentation result by optimizing the model in an iterative manner, and \cite{liu2020part} improved the performance by defining the prototype as a set of part-prototypes so that parts of an object can be recognized. However, prototype-based methods are limited in that the resolution of the predicted segmentation is often low since the relationship between the prototypes or parameters for prediction is learned in the down-sampled embedding space and then rapidly up-sampled using interpolation.

Recently, \cite{li2020fss} progressively performed upsampling of encoded features using a decoder with skip connections at different levels. In \cite{hu2019attention}, segmentation was performed using features obtained at various stages of the encoder with an attention mechanism. Also, \cite{roy2020squeeze} proposed a few shot method for 3D CT organ segmentation using dense connections with squeeze and excitation blocks added between the modules for support and query data. However, most of the proposed methods have limitations in obtaining smooth segmentation for 3D organs because they estimate segmentation of the query image by only relying on support data without considering contextual information between adjacent slices. Unlike the existing methods, we propose a 3D few shot segmentation method that can consider the relationship between adjacent slices for more reliable predictions.

\subsection{Recurrent Neural Networks for 3D Medical Image Segmentation}
A number of deep learning based segmentation methods have been proposed for 3D medical image analysis. Although some architectures are fully implemented with 3D convolutions, many methods predict 3D segmentation based on 2D slices \cite{zhao2019edge,oktay2018attention,roth2015deeporgan} or 3D patches \cite{korez2016model,zhu20173d,luna20183d} and then perform aggregation due to the expensive computations of 3D operations and limited training data. Though such methods can learn complex tasks relatively well, the main disadvantage is the lack of global context information used for prediction. To alleviate this problem, several methods have been proposed to consider consistency between adjacent slices using recurrent neural networks (RNN). In this setting, final segmentation is achieved by obtaining predictions on adjacent slices via U-net or FCN, then perform updates on the feature maps through RNNs. For example, \cite{xu2019lstm} and \cite{cai2017improving} feed U-Net predicted segmentation maps to an LSTM module. On the other hand, \cite{chen2016combining}, \cite{cai2018pancreas}, and \cite{li2019mds} used a bidirectional LSTM. \cite{bai2018recurrent} integrated a convolution based bidirectional LSTM into U-Net in order to model the relationship of encoded features between adjacent slices. Though successful, previous methods require a large amount of labeled training data to learn a robust model. Instead, we propose to integrate an RNN in the few shot setting for segmentation. In this way, we alleviate the issue of low training data and learn robust features considering context between slices for several organ segmentation tasks.

\subsection{Transfer Learning in Few Shot Learning}
Transfer learning has been shown to improve model performance for different tasks by leveraging deep models already trained on larger datasets with related tasks or characteristics. Recently, several methods have been proposed to address fine tuning in the few shot setting with limited target support samples. For example, \cite{caelles2017one} proposed a fine tuning approach for 1-shot video object segmentation using the first frame. This method was also used in metric-based few shot segmentation \cite{shaban2017one} even though it was limited in application i.e., used to update only part of the model. In addition, optimization based few shot methods also employ fine-tuning in the intermediate stages of training. \cite{finn2017model} proposed to temporarily update the model using support data and minimize the loss for each task, in turn good performance is obtained with only a few model updates. \cite{sun2019meta} proposed to only update the fine tuning module during the fine-tuning stage by disentangling learning of general and transferable knowledge in the modules. Inspired by prior methods, we introduce a method to learn optimal parameters for the target task by using a fine tuning process that performs additional updates by randomly dividing support data in the $K$-shot setting.

\begin{figure*} [h]
\begin{center}
\includegraphics[width=1.0\linewidth] {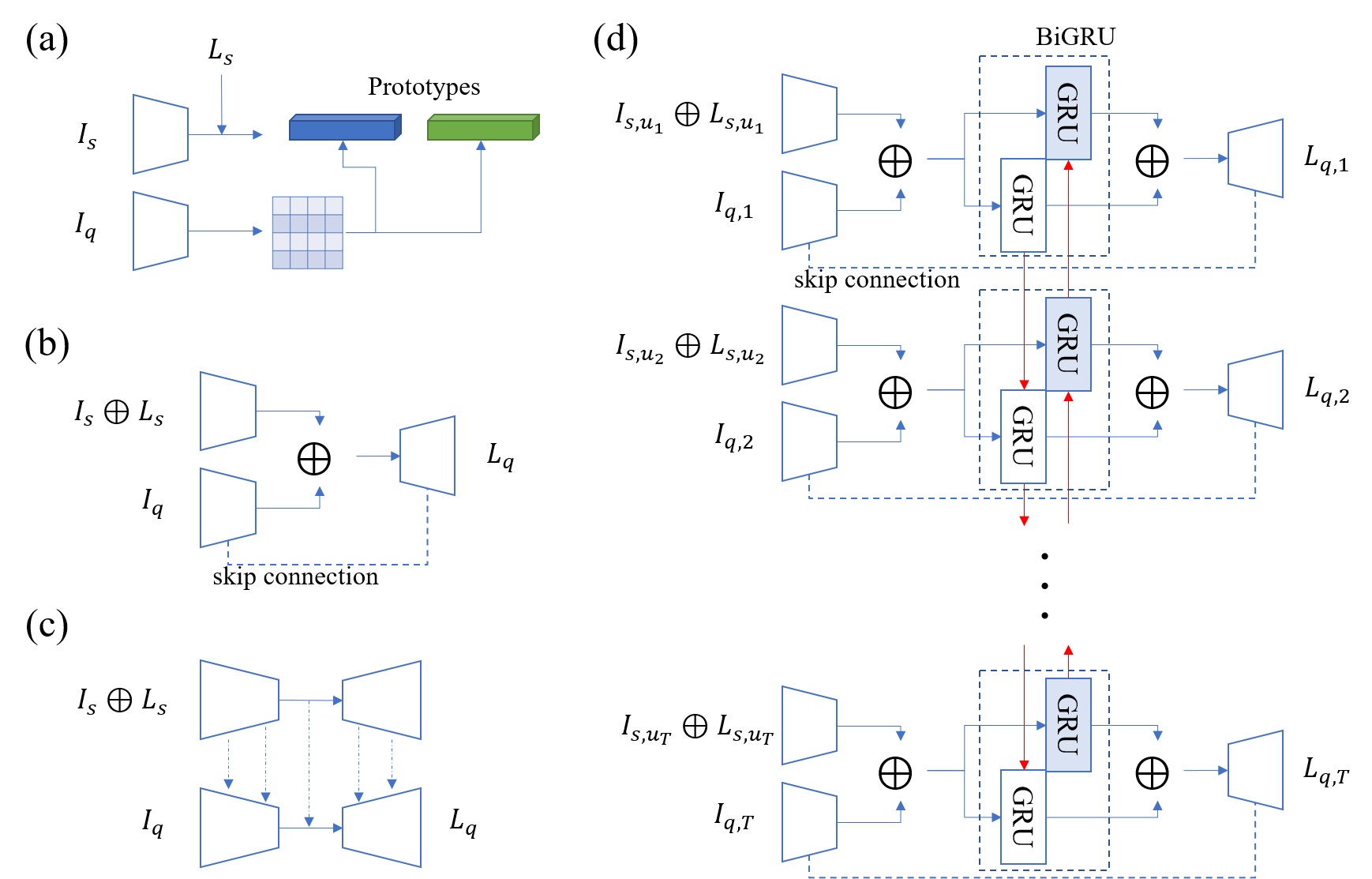}
\end{center}
\caption{Segmentation models using few shot learning. (a), (b), and (c) are the existing few shot segmentation models~\cite{wang2019panet,li2020fss,roy2020squeeze}, while (d) is our proposed method.}
\label{fig::overview}
\end{figure*}


\section{Methods}
\subsection{Problem Setup}
A few shot segmentation model $FSS_{\theta}$ learns parameters ${\theta}$ to segment a target object in a query image $I_{q}$ using $K$ pairs of support images and labels $\begin{Bmatrix}{I^1_{s}, L^1_{s}}\end{Bmatrix}$, $\begin{Bmatrix}{I^2_{s}, L^2_{s}}\end{Bmatrix}$, ..., $\begin{Bmatrix}{I^K_{s}, L^K_{s}}\end{Bmatrix}$, where $K$ defines the degree of supervision. Common architectures for 2D image based few shot segmentation are shown in Fig.~\ref{fig::overview} (a) to (c). (a) shows a prototypical network which defines prototypes for target object and background, and then performs segmentation using the distance to the defined prototypes, (b) is a relation network consisting of encoders and decoders for segmentation in a fully convolutional network, and (c) represents a network with dense connections between modules for support and query data. All of these methods learn the relations between support and query data of various segmentation tasks and utilize them to predict a target label in $I_{q}$ using $K$ support samples as:
\begin{equation}
\label{eq::base}
L_{q} = FSS_{\theta}(\begin{Bmatrix}{I^k_{s}, L^k_{s}}\end{Bmatrix}^{K}_{k=1}, I_{q}).
\end{equation} 


\cite{roy2020squeeze} extended this concept to 3D image few shot segmentation. They divided a query volume $\textbf{I}_{q}$ into multiple 2D slices $I_{q,1}$, $I_{q,2}$, ..., $I_{q,T}$ where $T$ is the number of slices in the axial view, and then segmented each query slice $I_{q,t}$ separately using the corresponding support slice $I_{s,u_t}$ and label $L_{s,u_t}$, where $u_t$ is the index of the support sample. To determine $I_{s,u_t}$ and $L_{s,u_t}$ in a 3D support volume $\textbf{I}_{s}$ with label $\textbf{L}_{s}$, it was assumed that the starting and ending slice locations of the organ of interest in $\textbf{I}_{q}$ and $\textbf{I}_{s}$ were known. The index of a support slice corresponding to $I_{q,t}$ is obtained via $u_t=round((t/T)\times \hat{T})$, where $\hat{T}$ is the number of interest slices in $\textbf{I}_{s}$. This assumption is reasonable since the organ of interest is in a similar position from person to person, e.g., the liver is always located in the upper right part of the abdomen, even though it varies in size and shape. In this setting, few shot segmentation of a 3D image can be represented by the following formulation:
\begin{equation}
\label{eq::base2}
\textbf{L}_{q} = \begin{Bmatrix}{L_{q,t}}\end{Bmatrix}^{T}_{t=1} = \begin{Bmatrix}{ FSS_{\theta}(\begin{Bmatrix}{I^k_{s,u_t},  L^k_{s,u_t}}\end{Bmatrix}^{K}_{k=1}, I_{q,t})}\end{Bmatrix}^{T}_{t=1}.
\end{equation} 
Most 2D based few shot segmentation models can follow this setting, but the relation between adjacent slices is not considered. In this study, we follow the mentioned problem setting but propose to incorporate adjacent slice information to accurately segment $I_{q,t}$. Formally,
	
\begin{equation}
\label{eq::our}
L_{q,t} =  FSS_{\theta}(\begin{Bmatrix}{ \begin{Bmatrix}{I^k_{s,u_t},  L^k_{s,u_t}}\end{Bmatrix}^{K}_{k=1}, I_{q,t}}\end{Bmatrix}^{t_0+n_a}_{t=t_0-n_a}),
\end{equation} 
where $2n_a+1$ is the number of adjacent slices and $t_0$ is the index of the center of multiple slices. To achieve this, we introduce a 3D few shot segmentation method using the bidirectional RNN.

\subsection{Bidirectional RNN based Few shot learning}
Our model performs segmentation in three stages; $(1)$ features of support and query images are extracted through two separate encoders $E_s$ and $E_q$, respectively. $(2)$ A bidirectional GRU models the relationship between features extracted from adjacent slices. $(3)$ Finally, using updated feature maps and low level features in $E_q$, the decoder predicts the segmentation. An overview of the one-shot segmentation model is shown in Fig.~\ref{fig::overview} (d).



\subsubsection{Feature Encoders}
We used two separate encoders $E_s$ and $E_q$ to extract features from support and query images since they receive inputs with different number of channels. $E_s$ receives 2-channel input which is a concatenation of $I_{s,u_t}$ and $L_{s,u_t}$, whereas $E_q$ receives 1-channel $I_{q,t}$ as input. We used an ImageNet VGG16 \cite{simonyan2014very} model in the encoder module given its robust feature extraction ability. Since VGG16 takes a 3-channel image as input, we randomly initialized the parameters of the first layer and concatenate the features from the two encoders as: 
\begin{equation}
\label{eq::enc}
x_t = E_{s}( \begin{Bmatrix}{ I_{s, u_t}, L_{s, u_t} } \end{Bmatrix}^{K}_{k=1} )\oplus E_{q}(I_{q, t}),
\end{equation}
then feed $x_t$ into the GRU model. Low-level features with different resolutions extracted from $E_q$ are used again in the subsequent stage.


\subsubsection{Bidirectional GRU}
After features $\begin{Bmatrix}{x_{t}}\end{Bmatrix}^{T}_{t=1}$ are extracted by encoders from $\begin{Bmatrix}{I_{s, u_t}, L_{s, u_t}, I_{q, t}}\end{Bmatrix}^{T}_{t=1}$, GRU models the change between adjacent slices. In particular, a bidirectional GRU has two modes i.e. both forward and backward directions for efficient feature representation. Features are sequentially feed into the forward GRU and later reversed for the backward model. Each GRU calculates two gate controllers $z_t$ and $r_t$ with $x_t$ and a prior hidden state $h_{t-1}$ for memory updates as: 
\begin{equation}
\label{eq::zt}
z_{t} = \sigma (conv_{z}(h_{t-1}\oplus x_{t})+b_{z}),
\end{equation}
\begin{equation}
\label{eq::rt}
r_{t} = \sigma (conv_{r}(h_{t-1}\oplus x_{t})+b_{r}).
\end{equation}
$z_t$ controls the input and output gates whereas $r_t$ determines which part of the memory would be reflected on the hidden state $h_t$. Formally,
\begin{equation}
\label{eq::hhatt}
\hat{h}_{t} = tanh (conv_{h}(r_{t} \odot h_{t-1}\oplus x_{t})+b_h),
\end{equation}
\begin{equation}
\label{eq::ht}
h_{t} = (1-z_{t}) \odot h_{t-1} + z \odot \hat{h}_{t}.
\end{equation}
In our GRU module, operations are replaced with $3\times3$ convolutions instead of weight multiplication in normal GRU cell. Sigmoid activation functions are used after the gate controller output, and a hyperbolic tangent function applied following the final hidden state output. Following, features extracted from the forward GRU $h^{f}_t$ and backward GRU model $h^{b}_t$ are concatenated as: 
\begin{equation}
\label{eq::htfin}
h^{biGRU}_{t} = h^{f}_t \oplus h^{b}_t,
\end{equation}
and then passed to the decoder. We used the GRU modules due to their low memory footprint, though any bidirectional RNN such as the Long Short-Term Memory (LSTM) model~\cite{hochreiter1997long} can alternatively be used for sharing features between adjacent slices.

\begin{figure} [h]
\begin{center}
\includegraphics[width=1.0\linewidth] {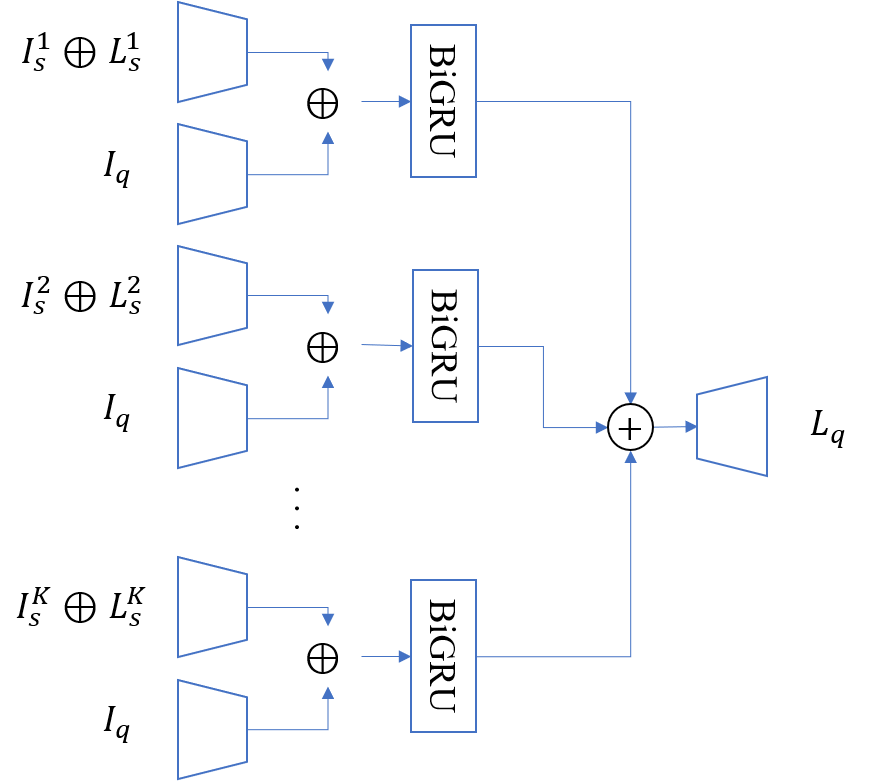}
\end{center}
\caption{Model architecture for $K$ shot segmentation. The features from bidirectional GRU are summed before the decoder.}
\label{fig::multishot}
\end{figure}

When $K$ support data $(I^{k}_{s}, L^{k}_{s})^{K}_{k=1}$ are used (see Fig.~\ref{fig::multishot}), the GRU operation is performed for each support and query data pair, then the obtained features are summed as: 
\begin{equation}
\label{eq::multishot}
h^{biGRU}_{t} = \sum^{K}_{k=1}h^{biGRU, k}_{t}.
\end{equation}
Finally, $h^{biGRU}_{t}$ is passed to the decoder.

\subsubsection{Decoder}
Our decoder has a similar architecture to UNet \cite{ronneberger2015u} for high resolution segmentation. The final segmentation is obtained by utilizing the features processed by the encoders and bidirectional GRU, as well as the low level features of the query slice image obtained from $E_q$. Low level features are connected to the decoder by skip connections, wherein the decoder predicts the segmentation using both low- and high-level information. Our cost function was defined as the summation of cross entropy loss and dice loss \cite{milletari2016v} between the prediction and label. 

\subsection{Transfer Learning based Adaptation}
Since the target organ is never seen in the training stage, it may be challenging for model to learn the optimal parameters for the target. Thus, in multi-shot scenarios, we perform transfer learning with few target-support samples prior to testing. Specifically, we temporarily sample support and query pairs from the support data and update our model. For example, in $K$ shot testing stage, we collect existing pairs of support and query data by choosing $K-1$ samples from the $K$ support set as temporary support samples and use the remainder as a temporary query image to retrain our model. Since we use $2n_a+1$ adjacent slices from a 3D volume, many different training pairs can be sampled enabling a robust fine tuning process. Moreover, to avoid overfitting and encourage training stability, we use random flipping and rotation based augmentation during training. Using this strategy, our model can effectively adapt to the novel characteristics of target data.



 




\begin{table*}[h]
\caption{Performance comparison of the proposed model $FSS_{BiGRU}$ against baseline models on the BCV dataset using the (dice score $\pm$ stdev) evaluation metric. The second column represents the number of training data ($\#$) and FT denotes the fine-tuning. Boldface represents the best accuracy among the few shot comparison methods. }
\begin{center}
\label{internal_test}
\newcolumntype{C}{>{\centering\arraybackslash}m{6em}}
\begin{tabular}{|c|c|C|C|C|C|c|}
\hline
Model & $\#$ & Spleen & Kidney & liver & bladder&mean \\
\hline
\hline
 U-net (lower) & 1  & 0.695 & 0.793 & 0.645 & 0.574 & 0.677\\
\hline
 U-net (upper)& 15  & 0.896 & 0.884 & 0.902 & 0.788& 0.867\\
\hline
\hline
 \multirow{3}*{$FSS_{base}$}&1& 0.722$\pm$0.010 & 0.846$\pm$0.017 & 0.751$\pm$0.027 & 0.515$\pm$0.045& 0.703  \\
\cline{2-7}
& 3& 0.781$\pm$0.033 & 0.829$\pm$0.021 & 0.788$\pm$0.008 & 0.496$\pm$0.038& 0.711  \\
\cline{2-7}
& 5& 0.829$\pm$0.015 & 0.861$\pm$0.008 & 0.849$\pm$0.008 & 0.585$\pm$0.008& 0.727  \\ 
\hline
 \multirow{3}*{$FSS_{prototype}$}&1& 0.758$\pm$0.049 & 0.759$\pm$0.033 & 0.788$\pm$0.014 & 0.584$\pm$0.204& 0.722  \\ 
\cline{2-7}
& 3& 0.834$\pm$0.018 & 0.791$\pm$0.006 & 0.836$\pm$0.011 & 0.696$\pm$0.016& 0.790  \\ 
\cline{2-7}
& 5& 0.832$\pm$0.012 & 0.787$\pm$0.016 & 0.843$\pm$0.004 & 0.701$\pm$0.009& 0.791  \\ 
\hline
$FSS_{SE}$ & 1 & 0.767$\pm$0.022 & 0.824$\pm$0.010 & 0.750$\pm$0.062 & 0.607$\pm$0.033& 0.737  \\ 
\hline
 \multirow{3}*{$FSS_{BiGRU}$}&1& 0.816$\pm$0.028 & 0.829$\pm$0.061 & 0.817$\pm$0.039 & 0.648$\pm$0.017& 0.778  \\ 
\cline{2-7}
& 3& 0.871$\pm$0.024 & 0.885$\pm$0.009 & 0.835$\pm$0.010 & 0.656$\pm$0.014& 0.812  \\ 
\cline{2-7}
& 5&0.888$\pm$0.013 & 0.889$\pm$0.005 & 0.864$\pm$0.006 & 0.698 $\pm$ 0.004 & 0.835 \\ 
\hline
$FSS_{BiGRU}$+FT& 5 & \textbf{0.905$\pm$0.019} & \textbf{0.900$\pm$0.006} & \textbf{0.887$\pm$0.007} & \textbf{0.771$\pm$0.030} & \textbf{0.866} \\ 
\hline
\end{tabular}
\end{center}
\end{table*}

\section{Experiments}

\subsection{Dataset}
The proposed method was evaluated on the Multi-atlas labeling Beyond the Cranial Vault (BCV) dataset \cite{landman2017multi} which was used in a challenge at MICCAI 2015. The dataset includes 30 3D CT scans with segmentation labels for 15 organs. Among 15 organs, the labels of 9 organs (spleen, left kidney, esophagus, liver, stomach, aorta, inferior vana cava, bladder, and uturus) were used in our experiments since the other 6 organs were too small or hard to segment even with a supervised learning method due to large shape variations. 

Moreover, we used two external datasets to see if the proposed model would be applicable to data with different characteristics. We employed CTORG \cite{CTORG2019} which contains 119 images with labels of 6 organs (lung, bones, liver, left and right kidneys, and bladder). It is worth noting that variations of in-plane resolution and thickness between images were significant since the dataset was collected from multiple sites. In our experiments, external tests were performed on the liver, kidney, and bladder, excluding the lung and brain since many CT scans did not include the whole part of the lung, and the brain samples were limited. Second, we also evaluated our method on the DECATHLON\cite{simpson2019large} dataset. It consists of several images with 10 different organs (CT Liver, multimodal MRI brain tumors, mono-modal MRI hippocampus, CT lung tumors, Multimodal prostate, monomodal left atrium, CT pancreas, CT colon cancer primaries, CT hepatic vessels, and CT spleen). In a similar fashion with earlier experiments, spleen and liver data were used in the external tests, excluding organs that were too small or had severe shape changes.


\begin{table*}[h]
\caption{Performance comparison on external datasets (dice score $\pm$ stdev). The U-net (BCV) is the model trained by BCV dataset, while the U-net (lower) and U-net (upper) are trained using the same external datasets. Due to different numbers of volumes for each organ, we indicate 5 numbers used as the number of training data ($\#$) for U-net(upper), e.g. 27 training volumes for spleen(DECATHLON). Boldface represents the best accuracy among the few shot comparison methods. }
\begin{center}
\label{ext_test}
\newcolumntype{C}{>{\centering\arraybackslash}m{6em}}
\begin{tabular}{|c|c|C|C|C|C|C|C|}
\hline
 & &  \multicolumn{2}{c|}{DECATHLON}&\multicolumn{3}{c|}{CTORG} \\
\hline
Model & $\#$ &Spleen &Liver & Kidney & liver & bladder  \\
\hline
\hline
 U-net (BCV) & 15 &  0.704 & 0.875 & 0.553 & 0.806 & 0.606 \\
\hline
 $FSS_{base}$ & 5 &0.848$\pm$0.017 & 0.802$\pm$0.009 & 0.657$\pm$0.015 & 0.779$\pm$0.003 & 0.522$\pm$0.013   \\ 
\hline
 $FSS_{prototype}$ & 5& 0.837$\pm$0.009 & 0.791$\pm$0.022 & 0.612$\pm$0.007 & 0.725$\pm$0.012 & 0.548$\pm$0.010  \\ 
\hline
 $FSS_{SE}$ & 1& 0.788$\pm$0.002 & 0.644$\pm$0.052 &  0.529$\pm$0.005 & 0.610$\pm$0.052 & 0.663$\pm$0.004 \\
\hline
 \multirow{3}*{$FSS_{BiGRU}$}&1 & 0.841$\pm$0.005 & 0.792$\pm$0.029& 0.623$\pm$0.031 & 0.751$\pm$0.015 & 0.703$\pm$0.009\\ 
\cline{2-7}
& 3&0.873$\pm$0.011 & 0.812$\pm$0.016 & 0.689$\pm$0.041 & 0.726$\pm$0.004 & 0.713$\pm$0.005  \\ 
\cline{2-7}
& 5& 0.878$\pm$0.029 & 0.852$\pm$0.013& \textbf{0.716$\pm$0.03} & \textbf{0.788$\pm$0.003} & 0.729$\pm$0.005 \\ 
\hline
$FSS_{BiGRU}$+FT & 5 & \textbf{0.900$\pm$0.022} &   \textbf{0.888$\pm$0.009} & 0.701$\pm$0.025 & 0.778$\pm$0.009 & \textbf{0.731$\pm$0.001} \\

\hline
\hline
 U-net (lower) & 1&0.571 & 0.625  & 0.621 & 0.727 & 0.243\\
\hline
 U-net (upper) & \footnotesize{27,87,65,63,53}& 0.858 & 0.927  & 0.915 & 0.867 & 0.767 \\
\hline

\end{tabular}
\end{center}
\end{table*}

%

\subsection{Experimental Settings}
The BCV dataset was divided into 15 volumes for training or selecting support data, 5 volumes for validation, and 10 volumes for testing for each organ.  In training, pairs of support and query data were randomly sampled from the 15 volumes with 8 organs except a certain target organ to train the few shot models. For testing, support data was randomly sampled among the 15 volumes for the target organ, while the 10 volumes were used as query images. Since performing experiments for all organs was time-consuming, we tested our model on 4 clinically important organs (spleen, liver, kidney, and bladder) that are not too small. For example, the adrenal gland was excluded because it appears in limited slices of CT scans and was often hard to figure out the organ's 3D structure. For external validation, the models trained on the BCV dataset were applied to 65 liver, 63 kidney, and 53 bladder samples in CTORG, and 27 spleen and 87 liver data in the DECATHLON dataset. Voxel intensities of all images were normalized to range from 0 to 1 and slices were cropped into squares with the same size for each organ, then resized to 256$\times$256. 

To show the effectiveness of proposed model, we compared our method with a U-net based supervised method and three few shot models ($FSS_{base}$ \cite{li2020fss}, $FSS_{prototype}$ \cite{wang2019panet}, and $FSS_{SE}$ \cite{roy2020squeeze}) recently proposed. A U-net \cite{ronneberger2015u} trained with only one sample per organ was used as a lowerbound while that trained with all accessible data was used as a upperbound model. For fair comparison with our proposed model, U-net was modified to use 5 adjacent axial slices as input and consisted of 2D convolutional encoder and decoder. He initialization \cite{he2015delving} was used for all models with Adam \cite{kingma2014adam} optimization and a learning rate of $10^{-4}$. For every iteration in the training stage, support and query volumes were randomly selected from training data containing various organ segmentation labels except the target organ. For the bidirectional GRU models, a total 5 slices were feed into the model, i.e., $n_a$ was set as 2. Similar settings were used for inference. In addition, the same parameter initialization and data augmentation (flipping and rotation) was applied across all evaluated models.


$FSS_{base}$ is a baseline model with a similar architecture to the proposed model if the bidirectional GRU module is omitted. $FSS_{prototype}$ uses prototypes and an alignment process for predictions. It defines prototypes of foreground and background to implement distance-based pixel-wise classification on reduced feature maps extracted by an encoder. On the other hand, $FSS_{SE}$ model uses squeeze and excitation blocks with skip connections trained from scratch, with a separate encoder and decoder for support and query data. Except for $FSS_{SE}$, we evaluated 1, 3, and 5 shot models on the internal and external testing datasets. Since $FSS_{SE}$ was designed for the one-shot setting, 3 and 5 shot settings were not considered for evaluation. As for the proposed model, we note it as $FSS_{BiGRU}$.


Since the performance of segmentation may vary depending on how the support set is selected, the experiment was performed with different support sets randomly sampled 5 times for each query sample, and we report the average across trials. Segmentation performance was measured by the dice similarity score between the prediction and label.

\begin{figure} [h]
\begin{center}
\includegraphics[width=1.0\linewidth] {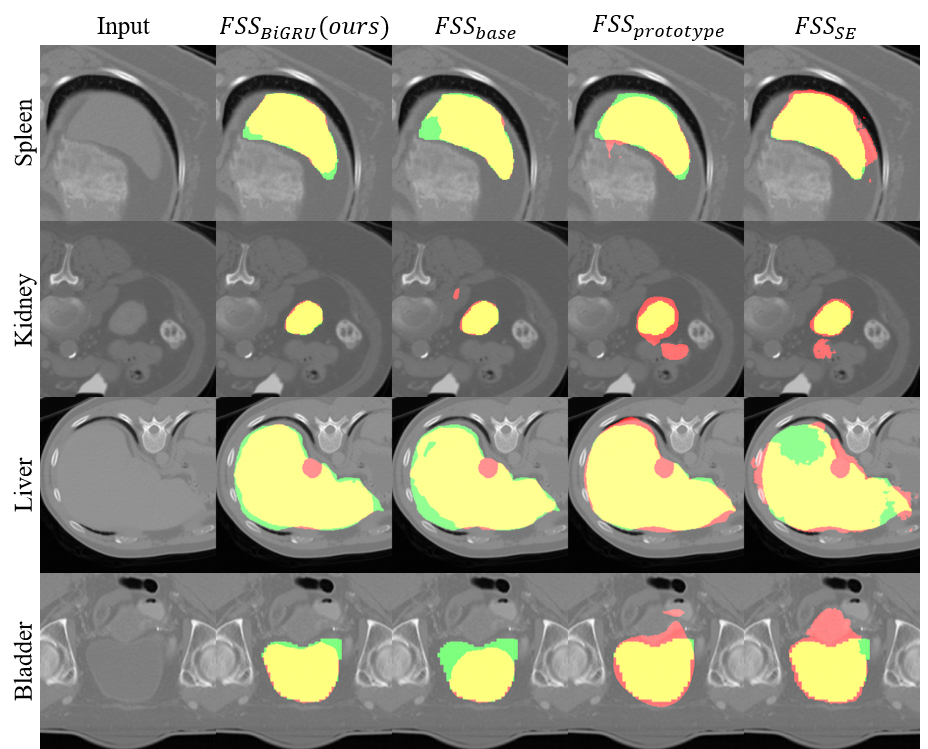}
\end{center}
\caption{Qualitative results in the axial view. Yellow denotes overlapping region (true positive), while green and red denote false negative and false positive regions, respectively.}
\label{fig::axial_view}
\end{figure}

\begin{figure} [h]
\begin{center}
\includegraphics[width=1.0\linewidth] {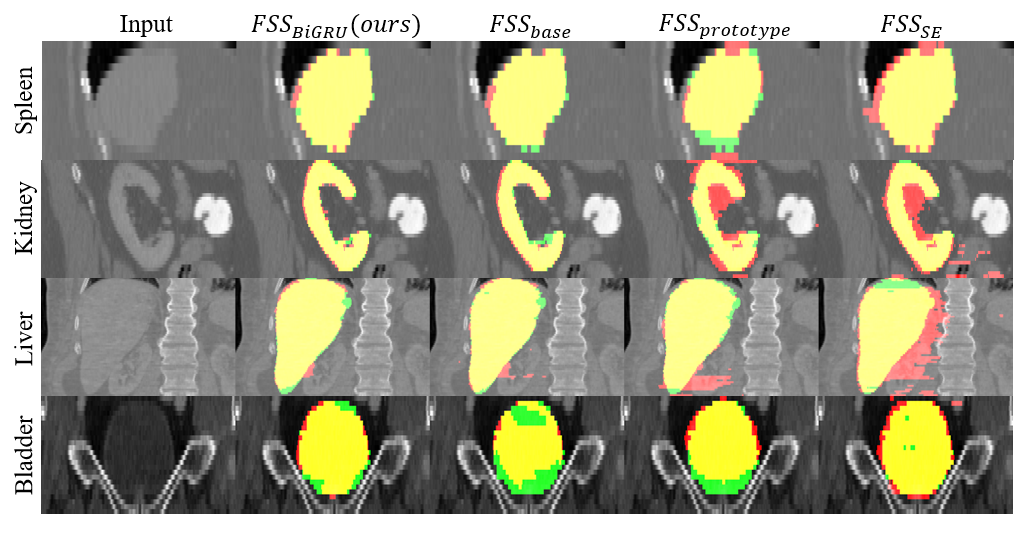}
\end{center}
\caption{Qualitative results in the coronal and sagittal view. Spleen and kidney are shown in the sagittal view, while liver and bladder are shown in the coronal view.}
\label{fig::side_view}
\end{figure}

\subsection{Results and Discussion}
\subsubsection{Internal Test}
We show an overall comparison for methods trained and tested on the BCV dataset in Table~\ref{internal_test}. $FSS_{BiGRU}$ with/without fine-tuning and its variants using different number of samples showed performance comparable to the fully supervised baseline. The margins was largely significant in the one-shot setting across all organs with an approximate 20\% mean score improvement. This clearly shows the proposed method is feasible for segmentation even under an extreme limited data regime.
	  
Notably, our method became better (i.e., the accuracy increased) and robust (i.e., the standard deviation decreased) as the data samples were increased in most cases. Though unsurprising that the upperbound had high scores on most organs, it is rather significant that our model showed similar performance. In addition, we noted that transfer learning largely improved overall performance i.e. +3\% mean score improvement over $FSS_{BiGRU}$. This implies that after additional updates our model was able to adapt the segmentation task of a target organ unseen in training. These results demonstrate that reliable segmentation can be achieved when considering multiple slices alongside 3D structural information to encode relationships between adjacent slices.




\subsubsection{External Test}

In Table~\ref{ext_test}, we further evaluate our approach on external datasets to assess model performance under distribution shift. For simplicity, we considered the 5-shot setting for $FSS_{base}$ and $FSS_{prototype}$. The upperbound trained on BCV as well as the upper- and lowerbound methods trained using all accessible data in the external set are also included for completeness. 

In general, we observed that the performance of upperbound models trained with BCV dataset deteriorated on the external datasets. The model achieved significantly reduced scores across most organs except the liver in DECATHLON dataset compared to the results in Table~\ref{internal_test}. This is due to the different scanning protocols and machinery employed in the clinical setup. It may be challenging to achieve reliable segmentation on external datasets whose resolution is different since the model may overfit to appearance of specific resolution. 

On the other hand, few shot learning based segmentation methods can alleviate this effect by capturing similarities between query and support samples in both training and testing. Notably, our model obtained the comparable performance with the internal test on the two organs in DECATHLON dataset and the bladder in CTORG dataset. Especially, with the transfer learning updates, we obtained improved performance in the DECATHLON dataset i.e. +3\% in both organs. This performance was comparable to the upperbound model in DECATHLON dataset. 

The performance of our method on the kideny and liver in CTORG dataset was significantly lower that that on the internal test. The performance of supervised learning was good in the case of CTORG dataset since there were relatively many training data. On the other hand, when there is no image with a resolution similar to that of the query image among the few support data, the performance of the few shot learning methods deteriorated. In this sense, the transfer learning strategy was not significant as well if the resolutions of support and query images are inconsistent. However, for this challenging task, the proposed method achieved the best performance among all the few shot learning methods. We believe that better results can be obtained if data with multiple resolutions are contained in the support set.



\subsubsection{Qualitative results}
Fig.\ref{fig::axial_view} and Fig.\ref{fig::side_view} show the qualitative results obtained in the axial view and other views, respectively. In most cases, the proposed method obtained similar segmentation to the ground truth label as opposed to the other few shot methods. Since the comparison methods do not consider information between adjacent slices, segmentation is often not smooth and false positives is obtained like noise in the outer part of the organ. This effect is more pronounced when the appearance of support and query images is different. It was confirmed that the difference in prediction between adjacent slices was relatively large in the sagittal or coronal views compared to the axial view in which the training was performed (see Fig. \ref{fig::side_view}). On the other hand, as the proposed method considers the information between adjacent slices together, the boundary appears smooth even in the sagittal and coronal views.


\section{Conclusion}
In this paper, we propose a novel framework for CT organ segmentation under a limited data regime. Our model reliably incorporates multi-slice information to achieve precise segmentation of unseen organs in CT scans. In addition, we show that a bidirectional RNN module can effectively model 3D spatial information for improved feature learning. To learn the optimal parameters for unseen target task, we further introduce a transfer learning strategy. Extensive evaluation on public datasets show the effectiveness and generalization ability of our proposed model. Our method achieved the segmentation performance comparable to the U-net based supervised learning model on internal and some of external datasets. For future work, we will develop few shot segmentation model that works on different body parts regardless of modality.




\bibliography{aaai21}
\end{document}